\pdfoutput=1

\documentclass[11pt]{article}

\usepackage{acl}

\usepackage{times}
\usepackage{latexsym}
\usepackage{graphicx}
\usepackage[T1]{fontenc}
\usepackage{amsfonts}
\usepackage{booktabs}

\usepackage[utf8]{inputenc}
\usepackage{colortbl}

\usepackage{microtype}
\usepackage{smile}
\usepackage{float}

\usepackage{inconsolata}
\usepackage{subfigure}

\newcommand{\ours}{\textsc{Arl2}}

\newcommand{\vpara}[1]{\vspace{1.5ex}\noindent\textbf{#1}}
\title{\ours{}: Aligning Retrievers for Black-box Large Language Models \\ via Self-guided Adaptive Relevance Labeling}

\author{Lingxi Zhang$^1$, Yue Yu$^2$, Kuan Wang$^2$, Chao Zhang$^2$ \\
 $^1$ Renmin University of China, Beijing, China \\
 $^2$ Georgia Institute of Technology, Atlanta, USA\\
  \texttt{zhanglingxi@ruc.edu.cn},~~\texttt{\{yueyu, kuanwang, chaozhang\}@gatech.edu} \\}
  
\begin{document}
\maketitle
\begin{abstract}

  Retrieval-augmented generation enhances large language models (LLMs) by incorporating relevant information from external knowledge sources.
  This enables LLMs to adapt to specific domains and mitigate hallucinations in knowledge-intensive tasks.
  However, existing retrievers are often misaligned with LLMs due to their separate training processes and the black-box nature of LLMs.
  To address this challenge, we propose \ours{}, a retriever learning technique that harnesses LLMs as labelers. \ours{} leverages LLMs to annotate and score relevant evidence, enabling learning the retriever from robust LLM supervision.
  Furthermore, \ours{} uses an adaptive self-training strategy for curating high-quality and diverse relevance data, which can effectively reduce the annotation cost.
  Extensive experiments demonstrate the effectiveness of \ours{}, achieving accuracy improvements of 5.4\% on NQ and 4.6\% on MMLU compared to the state-of-the-art methods.
  Additionally, \ours{} exhibits robust transfer learning capabilities and strong zero-shot generalization abilities\footnote{The implementation of \ours{} will be published at \url{https://github.com/zhanglingxi-cs/ARL2}.}.


\end{abstract}

\section{Introduction}

Retrieval-augmented generation (RAG) is a widely used technique for tailoring large language models (LLMs)
\citep{openai2023gpt4,palm2,touvron2023llama}
to specific domains and tasks.
By incorporating information from external knowledge sources, RAG enhances LLMs by prompting them with relevant evidence \cite{lewis2020retrieval,izacard2022few,shi2023replug}, without the need for expensive fine-tuning \citep{he2023unicron}.
These knowledge sources serve as a non-parametric reference, allowing LLMs to access up-to-date and customized corpora for answering questions.
RAG has shown promising results in improving LLM response accuracy for target tasks, while also helping to mitigate LLM hallucination \citep{ji2023survey}.

The practice of RAG for state-of-the-art LLMs often involves directly using standard retrievers (e.g. Google Search~\citep{lazaridou2022internet},
BM25~\citep{bm25}) or off-the-shelf dense retrievers (e.g., DPR~\cite{karpukhin2020dense}, Contriever~\cite{izacard2022unsupervised}) trained with supervised relevance signals.
However, the performance of these methods is limited by the mismatch between the retrieval and downstream tasks, as the retrieved \emph{similar} documents may not always be \emph{useful} for the queries despite their relevance.
In fact, retrieved documents with similar topics but irrelevant content may even mislead the LLM's predictions \citep{yu2023chain,shi2023large}.

To address the challenge of adapting retrievers for LLMs, several works propose joint training of retrievers and language models ~\citep{izacard2022few,lin2023ra,cheng2023lift}.
However, these methods require training the LLMs from scratch, which is impractical for cutting-edge LLMs due to their prohibitive training costs and black-box nature.
The recent RePlug method \cite{shi2023replug} offers a solution by refining the retriever for black-box LLMs.
RePlug utilizes language modeling scores of  the answers as a proxy signal to train the dense retriever.
However, such supervision for retriever training is indirect and may not be discriminative enough, especially when the questions could be directly answered through the parametric knowledge of the LLM.
Therefore, effectively adapting retrievers for black-box LLMs remains an unsolved challenge.

\begin{figure*}
  \centering
  \includegraphics[width=0.92\textwidth]{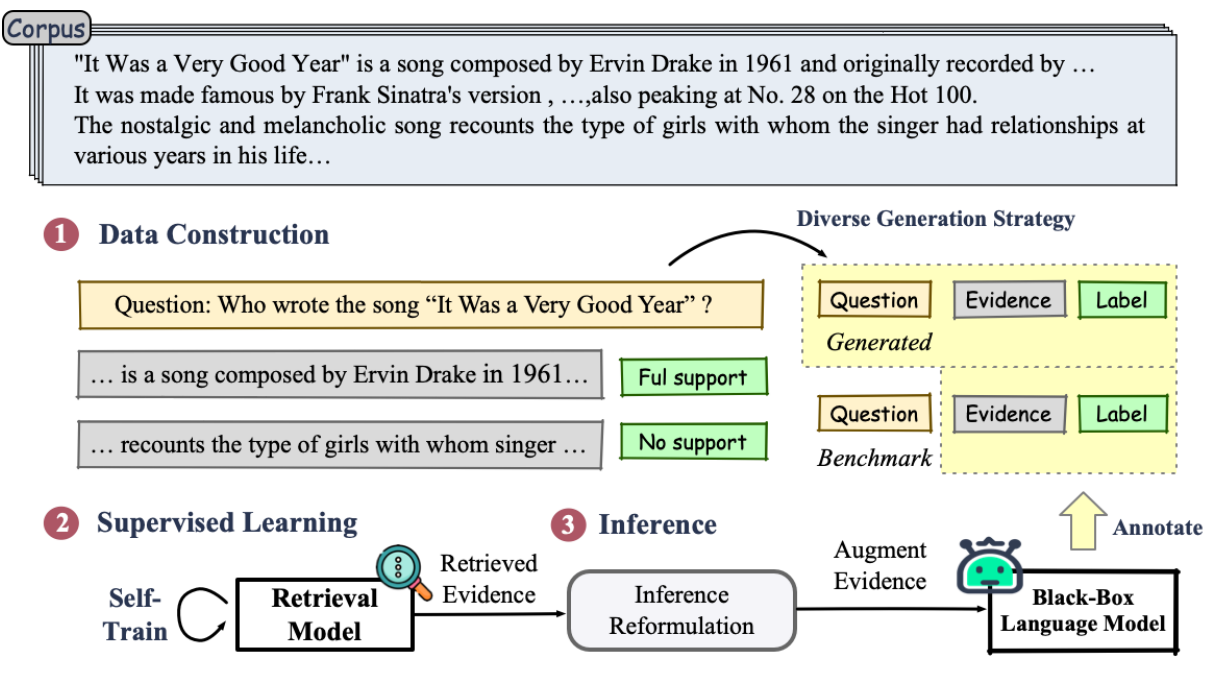}
  \caption{\textbf{Overview of \ours{}.} We first construct a diverse and high-quality training set of relevancelabelsl through LLM itself (Step 1), then we train the retriever with such relevance supervision (Step 2), and  finally, weinfere the LLM to yield answers through the reformulated augmented documents (Step 3).
  }
  \label{Fig.main2}
\end{figure*}

To enhance the retrieval model's performance, we present our approach, \ours{}\footnote{Short for \textbf{A}ligning \textbf{R}etrievers with \textbf{L}arge Language Models via Self-guided \textbf{A}daptive \textbf{R}elevance \textbf{L}abeling.}, which leverages guidance from LLMs through self-guided adaptive relevance labeling. Unlike existing methods that rely on indirect supervision via attention or answer-based language modeling scores, \ours{} leverages the LLMs' capabilities to directly assess document relevance, resulting in the curation of high-quality relevance labels to train better retrievers.

The advantages of using LLM-annotated relevance labels for retriever training are threefold. Firstly, \ours{} can effectively distinguish truly useful documents from similar but irrelevant ones, providing valuable positives and hard negatives for training. Secondly, it enables the creation of diverse training data beyond a single target dataset, surpassing the limitations of methods like RePlug. Thirdly, \ours{} can reversely generate diverse questions from unlabeled documents, enhancing data diversity and facilitating effective generalization under challenging few-shot or zero-shot scenarios.

To reduce the cost of data curation due to frequent LLM calls for relevance annotation, we propose an adaptive self-training strategy.
This strategy empowers the retriever to identify and label confident and well-trained data points, reducing the reliance on costly LLM interactions.
Furthermore, we introduce a cluster-driven prompt demonstration metric to ensure the diversity and high quality of the constructed data.
This enables learning a strong retriever from a smaller amount of high-quality and diverse annotated data.

Our experiments encompass both open-domain QA datasets (NQ and TQA) and a specific-domain QA dataset (MMLU).
The results demonstrate that our framework significantly enhances the performance of both the retriever and RAG-based question answering, achieving an accuracy improvement of 5.4\% on NQ and 4.6\% on MMLU.
Moreover, with the diverse curated data and self-training strategy, our method exhibits strong transferability to specific domains with limited training data and delivers promising results even in the challenging zero-shot generalization setting.

In summary, our contributions are as follows: (1) We propose \ours{}, a retrieval augmentation framework that effectively aligns retrieval models with black-box LLMs. \ours{} leverages the LLM as a labeler to assign relevance scores with robust LLM self-guided supervision. (2) We incorporate a cluster-driven prompt demonstration metric to ensure the generation of high-quality data.
Additionally, we explore a self-training strategy for the retriever to reduce the computational cost of LLM calls. (3) Extensive experiments demonstrate that our retrieval augmentation framework not only improves the performance of LLMs across various question-answering tasks but also exhibits strong transfer and zero-shot generalization capabilities.


\section{Related Work}
\vspace{-1ex}
\paragraph{Dense Retrieval.}  
Earlier research has explored various ways to learn representations for text retrieval~\citep{deerwester1990indexing, huang2013learning,gillick2018end}. 
With the rise of pre-trained language models, several works 
have presented the BERT-based dual-encoder as dense retrievers~\cite{lee-etal-2019-latent,karpukhin2020dense,xiong2021approximate}. They  typically employ encoders to independently encode queries and documents into a dense space, calculating the similarity via vector dot-product or cosine similarity. To further enhance performance of dense retrieval models, one line of approaches focuses on developing retrieval-oriented pretraining techniques~\cite{izacard2022few,gao-callan-2021-condenser,gao-callan-2022-unsupervised,yu-etal-2022-coco,xiao-etal-2022-retromae,ni-etal-2022-large,lin-etal-2023-train}, and another line of approaches focus on improving the negative contrast loss~\cite{ren2021rocketqav2,zhang2022adversarial}.
Additionally, some models utilize LLM-generated queries to generate synthetic examples for improving retrieval~\cite{ma-etal-2021-zero, dai2023promptagator}, 
but these dense retrievers are trained separately from LLM and may not always align well with the LLM, potentially resulting in sub-optimal performance when directly applied to target tasks~\citep{lin2023ra}.
A concurrent work \citep{wang2023improving} leverages LLM to generate synthetic text pairs from scratch without considering target tasks, thus being different from our focus. 


\paragraph{Retrieval-Augmented LLMs.} 
RAG have been widely used for language modeling ~\citep{retro,ram-etal-2023-context}, question answering~\citep{lewis2020retrieval,izacard2022few,shi2023replug}, and domain adaptation~\citep{xu-etal-2023-retrieval,xu2023weakly,shi2023retrieval}.
To align retrievers with LLMs, most RAG methods integrate a pre-trained retriever with a generator and subsequently undergo an end-to-end fine-tuning process to effectively capture knowledge~\cite{lewis2020retrieval}.
Among them, Atlas~\cite{izacard2022few} leverages retrieved documents as latent variables and fine-tunes retrieval models with four designed loss. AAR~\cite{yu-etal-2023-augmentation} identifies the LLM's preferred documents through FiD cross-attention scores~\cite{izacard-grave-2021-leveraging}, and fine-tuning the retriever with hard negative sampling. 
However, these methods are inapplicable to  black-box LLM as they require accessing LLM  parameters. 
The only exception is RePlug~\cite{shi2023replug}, which conducts supervised training by evaluating the KL divergence between the probability distributions of the retrieved documents and LLM's likelihood. 
\vspace{-1ex}

\section{Methodology}

In \ours{}, we employ LLMs to explicitly label relevance scores between questions and evidence, thereby generating relevance supervision for training an LLM-aligned retriever. \ours{} addresses two key challenges: (1) How can we effectively utilize the LLM to construct a diverse and high-quality training set of relevance labels? (Section~\ref{sec:data_construction}) and (2) How can we train the retriever using the provided relevance supervision and further leverage the retriever to inform adaptive LLM annotation? An overview of our proposed \ours{} method is presented in Figure~\ref{Fig.main2}.

\subsection{Data Construction}
\label{sec:data_construction}

To collect labeled data for retriever learning, it is crucial to go beyond query-document relevance as in standard information retrieval~\citep{lee-etal-2019-latent}.
In fact, the question $q$ in RAG applications is often under-specified and requires deeper language understanding.
Motivated by this, we leverage LLMs to provide direct supervision signals on the \emph{usefulness} for each piece of evidence for the question.

Specifically, we create training tuples denoted as $\mathcal{T}={(q_i, d_i, e_i, s_i)}_{i=1}^{|\cT|}$, where the evidence $e$ is a text segment extracted from the document $d$, and the variable $s$ represents the level of support for the question $q$ based on the evidence $e$. The relevance score $s$ can take on three values: 0 for ``\emph{no support}'', 0.5 for ``\emph{partial support}'', and 1 for ``\emph{full support}''.

We construct the corpus $\cD$ by compiling various corpora, such as WikiPedia~\citep{vrandevcic2014wikidata} and MS MARCO~\citep{ms_marco}. For each document $d$ in $\cD$, we generate a training tuple through a three-step process: \emph{question generation}, \emph{evidence identification}, and \emph{evidence scoring}. This procedure yields a dataset of 100,000 annotated instances, denoted as $\mathcal{T}_g$. Additionally, we curate 140,000 data instances $\mathcal{T}_b$ from QA benchmarks. The combination of these two sets forms the complete dataset $\mathcal{T}$.

\subsubsection{Question Generation}
\label{sec:question_generation}

To avoid manually generating questions from a vast amount of documents, we employ ChatGPT (\texttt{gpt-3.5-turbo-0613})~\citep{sun-etal-2023-chatgpt} for question generation to generate pairs in the format of $(q, e)$ from documents in the corpus $d\in\cD$.
This process involves providing the LLM reader with a specific prompt, namely, ``\emph{Given only the information below, following the examples, ask a factual question that we can answer according to the given passage}'', along with demonstrations to steer in-context learning.

\noindent \textbf{Diverse Generation Strategy.}
The quality of generated questions heavily relies on the selected document and the provided question examples.
Often, the generated questions follow similar patterns as the examples given.
Here, the question pattern refers to the sentence structure of the questions, such as special questions, yes-no questions, and declarative sentences.
To enhance the quality of the constructed data, we employ a diverse selection strategy.
First, we select relevant questions $q_r$ from $\mathcal{T}_b$ that share the same or a similar domain with the target document.
Second, we cluster the questions in $\mathcal{T}_b$ based on their patterns and opt for multiple question patterns from different clusters to generate diverse questions.
To facilitate diverse and high-quality question generation, we mask the entity mentions in the questions to focus on their core structure.
Each masked question is then mapped into a sentence embedding and clustered using ISODATA \citep{ball1965isodata}, a variation of K-means.
This makes the generated questions not only diverse but also of high quality.

\subsubsection{Support Evidence Annotation}
\begin{figure}
\includegraphics[width=0.48\textwidth]{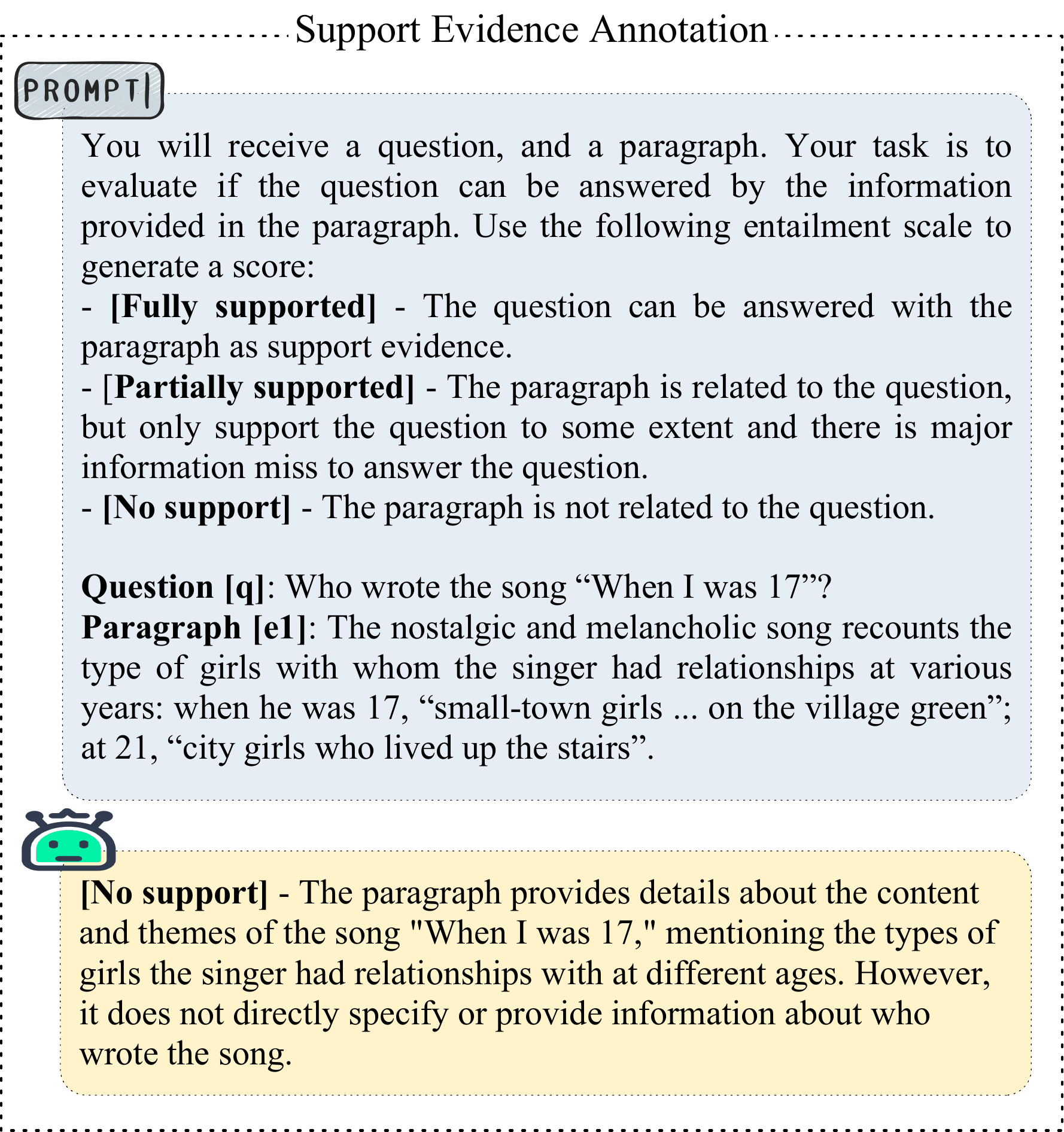} 
\caption{Illustration of prompting LLM to annotate relevance labeling data.}
\label{gpt_fig} 
\end{figure}

\noindent
\textbf{Evidence Identification.}
For each data point ($q_i$, $d_i$, $e_i$) in $\mathcal{T}_b$, we assign a support level label $s$ to indicate the relevance of the evidence $e_i$ to the question $q_i$. If ($q_i$, $d_i$, $e_i$) is directly obtained from human annotators, we set $s$ to \emph{``full support''}. Otherwise, we use ChatGPT as an annotator to obtain the support level label. Specifically, we provide ChatGPT with the question $q_i$, the document $d_i$, and an instruction to extract the supporting evidence from the document. ChatGPT then returns the extracted evidence and its relevance score to the question. 

\noindent
\textbf{Evidence Scoring.}
In addition to the positive samples with \emph{``full support''} labels, we also create negative samples for training the retrieval model. For each positive sample ($q_i$, $d_i$, $e_i$), we construct a challenging negative set $\mathcal{N}_i$ by selecting evidence segments from the same document $d_i$ or from other documents that are content- or domain-similar to $d_i$. We embed each evidence segment in $\mathcal{N}_i$ using SimCSE \cite{gao2021simcse}, a sentence embedding model. Then, we identify the top-$k$ most semantically similar evidence segments to $e_i$ based on their cosine similarity scores. Finally, we ask ChatGPT to label the relevance of these top-$k$ evidence segments to the question $q_i$. We discard the evidence segments labeled as \emph{``full support''} and retain the ones labeled as \emph{``partial support''} or \emph{``no support''} as negative samples.

\subsection{Retrieval Model Learning} 
\label{sec:retrieval_model}

In \ours{}, we train a dense retriever from the above instances of <question, evidence, relevance score>.
To enhance the learning of the retriever, we design pairwise and list-wise losses that incorporate hard negative sampling.
Furthermore, to reduce the cost associated with ChatGPT annotation, we propose a self-training strategy that enhances efficiency.

\subsubsection{Dense Retriever}
Given a query $q$ and an evidence corpus $\cD=\{d_1, d_2,\ldots,d_{|\cD|}\}$, the goal of the dense retriever is to first map all the documents $d\in\cD$ in a dense vector to build an index for retrieval,  then retrieve the top-$k$ most relevant documents via efficient vector similarity metrics.

We leverage the dual-encoder structure~\citep{lee-etal-2019-latent, karpukhin2020dense, lin-etal-2023-train} to embed query and documents using text encoders initialized from pretrained language models.
Specifically, the query encoder $E_Q(\cdot)$ and  document encoder $E_D(\cdot)$ map queries and documents $d\in\cD$ to low-dimensional real-valued vectors: 
\begin{equation}
\operatorname{sim}(q, d)=\cos\left(E_Q(q), E_D(d)\right).
\end{equation}

For efficient retrieval, we pre-compute the embedding of each document in $\cD$ and construct a FAISS index~\cite{johnson2019billion}.
\subsubsection{Learning Objective for Dense Retreivers}
To train the retriever, we employ a ranking loss. For each query $q_i$, we obtain the positive candidate $e^{+}$ along with a list of negative evidence candidates $\cN_{i}=\{e_1,e_2,...\}$. For each negative sample $e_j \in \cN_i$, we obtain its corresponding predicted relevance score $\hat{s}_{ij} = \operatorname{sim}(q_i, e_j)$ with the dual encoder and labeled relevance score $s_{ij}$ via LLM prompting (Figure~\ref{gpt_fig}). In specific, we optimize the retriever by
$$\mathcal{L} = \mathcal{L}_{\operatorname{list}} + \mathcal{L}_{\operatorname{pair}}, $$
where $\mathcal{L}_{\operatorname{list}}$ indicates a list-wise loss that discriminates between the positive and all listed negative evidence,
and $\mathcal{L}_{\operatorname{pair}}$ indicates a pairwise loss which focus on pairwise comparison between both (``\emph{full support}'' ,``\emph{partial support}'') and (``\emph{partial support}'',``\emph{not support}'').\\
\textbf{List-wise Contrastive Loss.} We utilize an InfoNCE loss which encourages positive instances to have high scores and negative instances to have low scores, which
is defined as:
$$\mathcal{L}_{\operatorname{list}}\left(s_i,\hat{s}_i\right)=-\frac{\exp \left(\hat{s}_i^{+} / \tau\right)}{\exp \left(\hat{s}_i^{+} / \tau\right)+\sum_{j=1}^{|\cN_i|} \exp \left(\hat{s}_{ij} / \tau\right)}$$
where $\hat{s}_i^{+}$ is the score for positive instance which is labeled as ``\emph{full support}'', and $\tau$ is a temperature parameter.  \\
\textbf{Pairwise Logistic Loss.} To better capturing the fine-grained relevance information beyond binary relevance labels, 
we further leverage a pairwise loss to deal with the ``\emph{partial support}'', which is defined as follow,
$$
\mathcal{L}_{\operatorname{pair}}\left({s}_i,\hat{s}_i\right)=\sum_{j=1}^{|\cN_i|} \sum_{j^{\prime}=1}^{|\cN_i|} \mathbb{I}_{s_{i j}>s_{i j^{\prime}}} \log \left(1+e^{\hat{s}_{i j^{\prime}}-\hat{s}_{i j}}\right)
$$
The aim of employing pairwise logistic loss is to ensure that a partially supported document can achieve a higher score than a fully negative one, while still scoring lower than a fully supported positive document.

\paragraph{Negative Sampling.}
Another important aspect of the above learning objective is how to mine negative examples $\cN_i$ for each query $q_i$. 
Here we propose a multi-step bootstrapped strategy to gradually provide ``\emph{harder}'' negative examples to effectively train the retriever.  
Initially, we kickstart the retriever training using in-batch negative samples as the 
warmup~\cite{gillick2018end,lee-etal-2019-latent}. Then, we choose the ``\emph{partial support}'' document as hard negative for both loss for further model training.  
This advanced negative sampling strategy significantly improves the retriever's performance compared to using randomly selected or BM25 selected negative samples~\citep{karpukhin2020dense}. 

\subsubsection{Adaptive Relevance Labeling}  
Although data collection can proceed without human annotations, utilizing ChatGPT series APIs comes with associated costs. To mitigate these expenses, we propose an adaptive strategy wherein we initially label only a subset of support evidence training data $\mathcal{T}_s$ first~\citep{li-etal-2023-coannotating,yu-etal-2023-cold}. After the warm-up epochs trained by $\mathcal{T}_s$, given a ($q_i$,$d$) pair without annotation, we segment document $d$ into text chunks $d = \{e_{i,1},...,e_{i,n}\}$
and enable the retriever to select the support evidence $e_{ij}$ with highest relevance score as
$$
s(q_i) = \max_{j\in\{1,2,\ldots,n\}}\operatorname{sim}(q_i, e_{i,j}).
$$
Such a confidence score $s(q_i)$ reflects the model's confidence on question $q_i$. 
Rather than evaluating confidence for each question individually, we first cluster questions (following the same pipeline as in Sec.~\ref{sec:question_generation}) and calculate the confidence score for each cluster 
$C_1,C_2,..C_m$, which denotes as $$s(C_i) = \sum_{q_t \in C_i} s(q_t) / |C_i|.$$
When a cluster exhibits a high model confidence score, we depend on the retriever's predictions and integrate all confident predictions within the cluster as supplementary training data. Specifically, we fine-tune the retriever using both the original training data and the newly generated training data in the subsequent epoch.
Conversely, if a cluster's confidence score is low, we resort to the data construction process outlined in Sec.~\ref{sec:data_construction}, which utilizes ChatGPT to create more relevance labels.

\subsection{Inference}
Formally, the LLM conditions on both the input question $q$ and the support evidence $\cD_q$ to generate a textual output $y$ as the answer. To augment the support evidence, a simple way is 
to input the evidence in $D_q$ from the highest relevant to the least relevant. However, the LLM is shown to be not robustly accessing or using information in long input contexts and may lose information in the middle~\citep{liu2023lost}. Therefore, we reorder the top-$k$ documents,
placing the most relevant document at either the beginning or the end of its input context.  Concretely, we cut the $D_q$ into three subsets according to the relevance score and put $d_1$ to $d_j$ at the beginning, $d_{j+1}$ to $d_{2j}$ at the end, and the rest at the middle, showing as following 
$$R_1 = d_1\circ...\circ d_j\circ d_{2j+1}\circ...\circ d_{k} \circ d_{2j} \circ ...\circ d_{j+1}$$
where the $\circ$ denotes the concatenation of two sequences.
To ensure robustness, we further rearrange the order 
within each subset for $N$ times to obtain permutations $\cR=\{R_1, \ldots, R_n\}$, and then ensemble the likelihood provided by the LLM for each $R_i$ to calculate the final answer score as
$$p(y | q, D_q) = \sum_{i=1}^{N}{P(y | q \circ R_i)}.$$

\section{Experiments}
\subsection{Experimental Setting}
\textbf{Datasets.} We evaluate the effectiveness of our framework through various open domain QA datasets, encompassing both general tasks including Natural Questions (NQ)~\citep{nq}, TriviaQA (TQA)~\cite{joshi-etal-2017-triviaqa} as well as domain-specific tasks like MMLU~\citep{mmlu}. 
NQ comprises natural questions extracted from real Google search, while TQA emphasizes trivia questions.
MMLU is a multiple-choice QA benchmark that can be categorized into humanities, STEM, social sciences, and others. 
For the document corpus, we follow DPR~\cite{karpukhin2020dense}, and construct a corpus of 300,000 QA pairs in total, which consists of 8.8 million passages from MS MARCO~\cite{ms_marco} and about 21 million passages from Wiki dump~\cite{vrandevcic2014wikidata}.
The size of each dataset is detailed in Table~\ref{question_num}.\\
\textbf{Baselines.} We compare our methods with strong baselines for open-domain question-answering tasks, which can be divided into: traditional joint training approaches and LLM-based approaches. Specifically, traditional joint training approaches encompass RETRO~\citep{retro}, UnifiedQA~\cite{khashabi-etal-2020-unifiedqa}, and Atlas~\citep{izacard2022few}. The LLM-based approach involves prompting the LLM to generate answers to questions. Representative LLM examples in our evaluation include Chinchilla~\cite{chinchilla}, PaLM~\cite{palm2}, ChatGPT~\cite{openai2023gpt4}, Codex~\cite{codex}, GenRead~\cite{yu2023generate}. To establish stronger baselines, besides raw LLMs, we compare our methods with retrieval augment methods which incorporate well-trained retrievers such as Contriever~\cite{izacard2022unsupervised}, Replug~\cite{shi2023replug}, both in a few-shot training setting, and also we evaluate the fully supervised DPR~\cite{karpukhin2020dense}.\\
\textbf{Evaluation Metrics.}
We evaluate the Recall@Top-k (k = 10, 20) for retriever performance, measuring whether the answer matches a text span within the Top-k retrieved passages.  For the overall precision, we consider accuracy (Acc.), verifying whether the answer generated by the RALM exactly matches the ground true answer in the dataset.\\
\textbf{Settings.} Based on DPR, we trained \ours{} with generated training instances based on the above corpus along with the benchmark dataset. Additionally, we evaluate different versions of \ours{}. \textbf{\ours{} (few-shot)} is a few-shot variant which trained on contriever and solely on generated instances along with few-shot benchmark training data. Furthermore, we examine \textbf{\ours{} (w/ re-ranker)}, an enhanced version that utilizes a generation-based re-ranker to improve retriever performance. Specifically, we employ the retriever to fetch the top 50 passages for each question and segment each document into text chunks for re-ranking, which ensuring to be within 50 words or 3 sentences. The re-ranker is trained with the same loss and training instances as the retriever. We evaluate all versions of our framework on ChatGPT with 20-shots.
\begin{table}[t]
\centering
\caption{Number of questions in each QA dataset.}
\label{question_num}
\scalebox{0.76}{
\begin{tabular}{lcccc}
\toprule
Dataset &  Train  & Dev & Test & Task \\
\hline NQ & 79,168  & 8,757 & 3,610 & Open-Domain \\
TriviaQA & 78,785  & 8,837 & 11,313 & Open-Domain \\
MMLU & 15,908  & 1,540 & 14,079 & Multi-Choice \\
MS MARCO & 1,010,916  & -- & -- & Retrieval \\
\bottomrule
\end{tabular}}
\end{table}
\begin{table*}[]
\caption{Overall Accuracy on QA datasets (\%).}
\centering
\renewcommand\arraystretch{0.97}
\label{tb:overall_result}
\scalebox{0.8}{
\begin{tabular}{l|c|c|ccccc}
\toprule
& \bf Natural Question & \bf TriviaQA & \multicolumn{5}{c}{\bf MMLU}          \\
\cmidrule(lr){2-2} \cmidrule(lr){3-3} \cmidrule(lr){4-8}
Model                    & All             & All     & All  & Hum. & Soc. & STEM & Other \\ \midrule
Chinchilla~\citep{chinchilla}              & 35.3            & 64.7    & 67.5 & 63.6 & 79.3 & 55.0 & 73.9  \\
PaLM~\citep{palm2}                     & 39.6            &  --  &  69.3    &   \underline{77.0}   &   \underline{81.0}   &  55.6    &   69.6    \\
Codex~\citep{codex}                    & 40.6            & 73.6    & 68.3 &   74.2   &  76.9    &   57.8   &   70.1    \\
ChatGPT~\citep{openai2023gpt4}                  & 38.7            & 74.2    & 70.3 &  75.1    &    76.8  &   59.3   &    70.8   \\ 
GenRead~\citep{yu2023generate} & 54.0 & 74.3 &  --   &  --   &  --   &  --   & -- \\
\hline
RETRO~\citep{retro}                    & 45.5            & 69.9    &   --   &  --    &   --   &  --    &    --   \\
UnifiedQA~\citep{khashabi-etal-2020-unifiedqa}                    & 55.9            &   --   &   48.9   &   45.6   &  56.6    &  40.2    &   54.6    \\
Atlas~\citep{izacard2022few}                    & 60.4            & 79.8    & 66.0 &   61.1   &   77.2   &  53.2    &    74.4   \\ \hline
ChatGPT+Contriever~\citep{izacard2022unsupervised}           & 44.2            & 76.0    & 69.9 &   68.1   &   76.6   &   50.8   &   74.8    \\
ChatGPT+DPR~\citep{karpukhin2020dense}                  & 58.0            & 76.9    & 72.9 &   69.6   &    80.6  &   64.2   &  78.6     \\
ChatGPT+Replug~\citep{shi2023replug}               & 45.4            & 77.8    & 71.8 &   76.5   &   79.9   &   58.9   &  73.2     \\
\rowcolor{violet!8}  ChatGPT+\ours{} & \underline{62.3}            & \underline{82.4}    & \underline{75.7} &  73.2    &    80.9  &  65.5    &    \underline{80.1}   \\
\rowcolor{violet!8}  ChatGPT+\ours{}~(few-shot)     & 54.9            & 81.0    & 73.9 &  70.8    &   80.4   &  \underline{66.7}    &   78.8    \\
\rowcolor{violet!15}  ChatGPT+\ours{}~(w/ re-ranker)  & \textbf{65.9}            & \textbf{85.9}    & \textbf{76.4} & \textbf{78.3} & \textbf{83.2} & \textbf{68.0} & \textbf{82.7}  \\ \bottomrule
\end{tabular}}
\end{table*}
\subsection{Overall Result}
The overall results are shown in Table~\ref{tb:overall_result},with bold indicating the best and underline indicating the second-best performances. \\
\textbf{Our retriever can enhance LLMs: } Both \emph{``\ours{}''} and \emph{``\ours{} (w/ re-ranker)''} outperform raw ChatGPT demonstrating that the retrieved evidence can correct the factual errors within ChatGPT's in-parameter knowledge. Especially, we exhibit a remarkable 6\% gain on MMLU, as MMLU involves extensive knowledge in a specific domain and certain numerical statistics that are often messed in ChatGPT's in-parameters. Our retrieved evidence contains the correct statistics, which can assist ChatGPT in correcting its answers. \\
\textbf{\ours{} surpasses other retrieval augmentation methods}, including strong baseline RePlug and other traditional retrievers. These outcomes suggest that our model adapts more effectively to LLM because it's trained on LLM-labeled data, ensuring that the retrieved evidence aligns better with LLM requirements. 
Additionally, we compare our model with some joint training baselines (RETRO, Atlas and UnifiedQA), owing to the strong capabilities of ChatGPT, augmenting black-box LLM with retrieved evidence effectively reduces the issue of hallucinations and surpasses joint training of medium-size PLM.\\
\textbf{Our re-ranker significantly improves retrieval performance}, as indicated in the table. \emph{``\ours{} (w/ re-ranker)''} outperforms pure retriever baselines. The re-ranker divides passages into multiple pieces of evidence, enabling a fine-grained ranking with our adaptive labeling strategy. Because the evidence is shorter than the entire passage and exhibits higher recall with a stronger base model, the re-ranker provides a more concise and accurate augmentation for the language model. This is particularly advantageous due to input length limitations imposed by the language model, resulting in a much better performance compared to \emph{``\ours''}.
\begin{table}[]
\caption{Performance on few/zero-shot transfers (\%).}
\label{tb:few-shot}
\centering
\renewcommand\arraystretch{0.95}
\scalebox{0.67}{
\begin{tabular}{l|ccc|cc}
\toprule
     & \multicolumn{3}{c|}{MMLU (Few-shot)}      & \multicolumn{2}{c}{TQA (Zero-shot)} \\
    \cmidrule(lr){2-4} \cmidrule(lr){5-6} 
 & Soc. & Hum. & STEM & All & R@20  \\ \midrule
PaLM  & 81.0  & 77.0 & 55.6 & -- & -- \\
Atlas & 54.6 & 46.1 & 52.8 & 79.8 & --\\
ChatGPT  & 75.8  & 73.4  & 69.9 & 74.2 & -- \\ \midrule
ChatGPT+Replug  & 79.9  & 76.0   & 72.1 & 77.8 & 76.2 \\
ChatGPT+Contriver & 80.3           & 74.1  & 71.9 & 76.0 & 74.2 \\
\rowcolor{violet!8}  ChatGPT+\ours  & \textbf{81.7} & \textbf{79.2}       & \textbf{74.4} & \textbf{77.9} & \textbf{77.2}\\ \bottomrule
\end{tabular}}
\end{table}

\subsection{Generalization and Transfer Ability}
\textbf{Few-shot Transfer Performance.}
We first evaluate the few-shot transfer ability of our retriever. In our training process, we solely pre-trained the retriever using data derived from or constructed from the MS MARCO datasets which only contain document corpus and questions, following Contriever. Subsequently, we employed few-shot questions (setting as 20 in our experiment) from MMLU as in-context learning examples to generate additional training data exclusively based on the document corpus. 
As the result shown in the left two columns of Table~\ref{tb:few-shot}, our model has surpassed both the raw LLM and other baseline retrieval methods, indicating its superior transfer capabilities. 
As our model fully utilizes the few-shot examples, generating training data patterns similar to those in the MMLU dataset, our retriever is familiar with the setting in MMLU even with a few data.  

\vpara{Zero-shot Generalization Performance.} We evaluate the zero-shot ability by training a retriever solely on the benchmark data in the NQ dataset and the constructed data from the NQ corpus, then directly test the model on TQA dataset. As shown in right column of Table~\ref{tb:few-shot}, the result demonstrates that our retriever has stronger generalization abilities. Additionally, we provide the retriever's recall rate to emphasize that our model's transfer ability isn't solely attributable to ChatGPT's robust generalization capabilities, the diverse select strategy has imparted a level of generalization to the retriever itself by maintaining a diverse range of training data across various domains and patterns.
\subsection{Ablation Study}
As shown in Table~\ref{tb:ablation_study}, we here perform the ablation study to demonstrate the impact of each strategy in construct relevance label and training the retriever. 
\begin{table}[!htbp]
    \centering
    \renewcommand\arraystretch{0.9}
    \caption{Ablation study on NQ and MMLU. (\%) }
    \label{tb:ablation_study}
\scalebox{0.90}{
\begin{tabular}{l|cc|c}
\toprule
    Methods   & \multicolumn{2}{c}{NQ}      & MMLU \\ \cmidrule(lr){2-4} 
    & Acc & R@20  & Acc   \\ \midrule
    \ours & 62.3  &84.3 & 75.7   \\ \midrule
    \multicolumn{4}{l}{\emph{Effect of Learning Objective}} \\ \midrule
    w/o pairwise & 59.0  &82.1 & 75.2  \\
    w/o listwise & 58.5  &81.2 & 74.8  \\
    w/o neg sample & 62.1  & 77.0 & 73.3 \\ \midrule
    \multicolumn{4}{l}{\emph{Effect of Relevance Labeling}} \\ \midrule
    w/o partial  & 55.4  &77.3 & 69.5  \\
    w/o label data & 48.7  &68.5 & 68.2 \\ \midrule
    \multicolumn{4}{l}{\emph{Effect of Inference Reformulation}} \\  \midrule
    w/o ensemble & 59.7  &-- & 75.6  \\
    w/o middle-rank & 59.1  &-- & 75.2 \\   \bottomrule
    \end{tabular}}
\end{table}

\vpara{Impact of Ranking Loss.}
Our final model leverage both pairwise and list-wise, and we here perform three models with different loss setting, \emph{``w/o pairwise''} indicates training the retriever without the pairwise loss, \emph{``w/o listwise''} indicates training the retriever without the list-wise loss.
The results show that remove pairwise loss and list-wise loss will drop recall, and leveraging both loss can achieve a better performance, as it both consider multiple random simple negative evidence to warm up but also take consider hard negative evidence through the pairwise to further improve models' performance. 
We also evaluate \emph{``w/ neg sample''} which replace our negative sample strategy with BM25 negatives~\citep{karpukhin2020dense}, and the results show the efficient of our strategy by bring harder cases for model training.

\begin{figure}[!htbp]
\centering
\setlength{\abovecaptionskip}{0cm}
\includegraphics[width=0.40\textwidth]{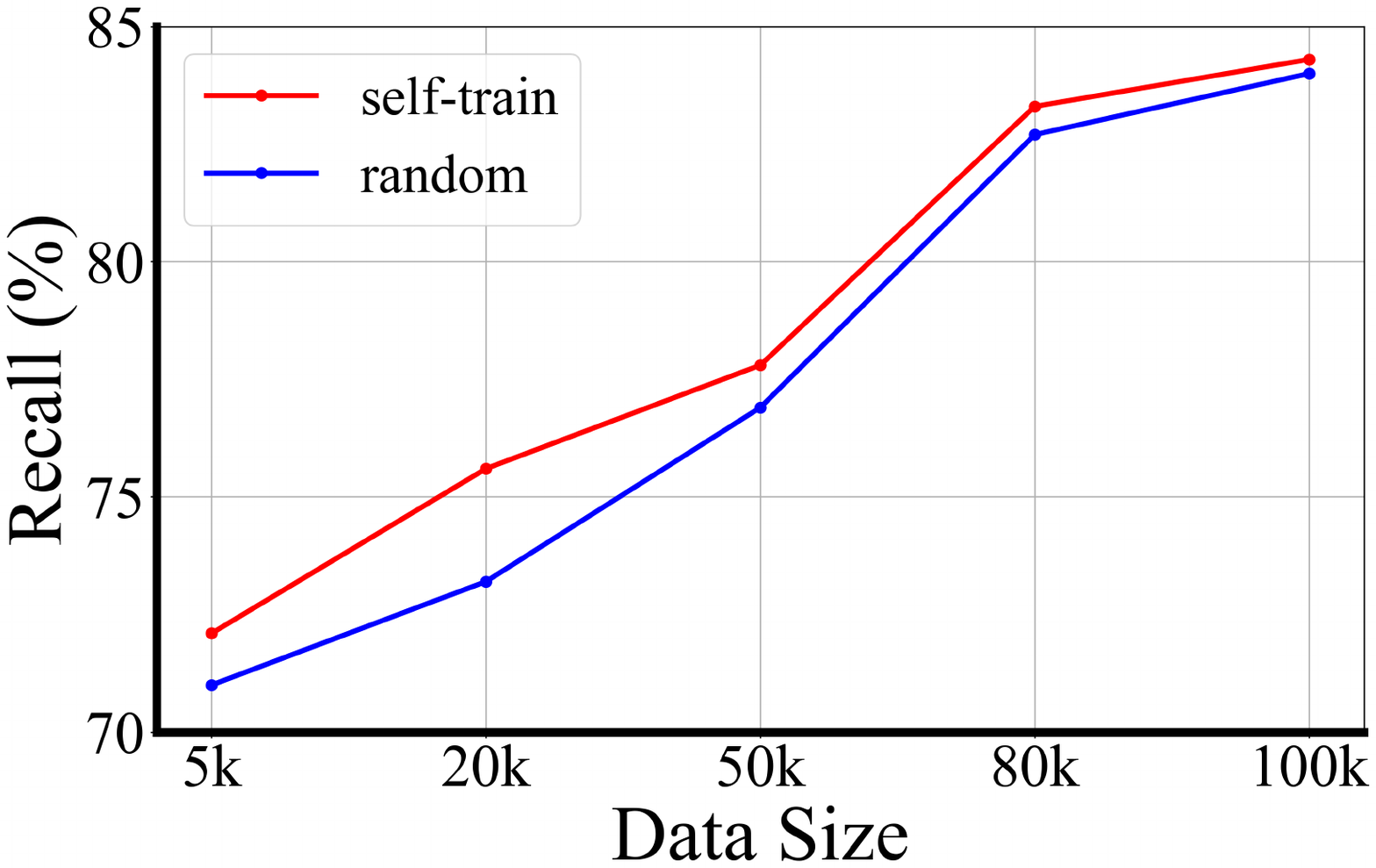} 
\vspace{-1ex}
\caption{Effect of annotated data size on NQ.}
\vspace{-1.5ex}
\label{exp_fig} 
\end{figure}
\vpara{Impact of Relevance Label Training Instance.}
Our labeled data can contribute the retriever's performance mainly on two aspect, the labeling of partial support evidence and the construction of more training data. So we perform ablation study on these two aspect, where \emph{``w/o partial''} denotes removing all partial label, and \emph{``w/o label data''} denotes removing all annotated data. The results show that both these removing drop accuracy, but \emph{``w/o label data''} drop rapidly, which indicates that more data is  more important than detailed training data. 

\vpara{Impact of Less Training Data.}
Considering the costly nature of API calls for data generation, 
we evaluate model performance with a smaller subset as training data. 
As shown in Figure~\ref{exp_fig}, the result illustrates that increased training data leads to enhanced performance, with optimal results achieved when all generated data are utilized. However, the rate of improvement declines with the increasing of training data, 
indicating that by selectively labeling diverse data, costs can be reduced while maintaining comparable performance. Compare to the diverse self-train strategy, we also show the performance of randomly selection and training solely based on it. As shown, the self-train method can save the cost and achieve better result within same less data, which is more efficient.

\vpara{Impact of Inference Reformulation.}
For inference, analysis the impact of two input reformulation strategies, \emph{``w/o ensemble''} indicates the model directly takes the result from middle-rank input, \emph{``w/o middle-rank''}  indicates the model take a simple positive rank input. And the result shows that both these two strategy can benefit LLM's performance. The accuracy does not drop much indicates that although model may lost information in middle, LLM can still obtain some external factual information from the input augment data.

\section{Conclusion}
We introduce \ours, a retrieval-augmentation framework that harnesses the capabilities of LLMs as annotators for retriever learning. Unlike conventional approaches that face misalignment due to separate training processes and the inherent complexity of LLMs, our framework dynamically leverages LLMs to annotate and assess relevant evidence. This enables the retriever to benefit from robust LLM supervision. Additionally, we incorporate a self-training strategy to mitigate the cost associated with API calls.
Through extensive experimentation, we demonstrate the effectiveness of \ours, which enhances accuracy in open-domain QA tasks, exhibits robust transfer learning capabilities, and showcases strong zero-shot generalization abilities.

\section*{Acknowledgements}
We would like to thank reviewers from the ACL Rolling Review for the helpful feedback.  
This work was supported in part by NSF IIS-2008334 and CAREER IIS-2144338.

\section*{Limitations}
In this research, annotating relevance labels can be expensive due to the extensive use of ChatGPT APIs. Our future work will further explore strategies to generate diverse, high-quality data to reduce these costs. Additionally, we aim to expand the curated relevance data to cover more specific domains like biomedical and life sciences. Correspondingly, we will evaluate the method's performance on tasks from such domains to assess its generalizability.

\bibliography{anthology,custom}

\appendix
\section{Appendix}
\label{sec:appendix}
\subsection{Implementation Details}
To enhance the efficiency of the training process, we establish a FAISS index for rapid similarity searches. To train the model, we employ the AdamW optimizer~\cite {loshchilov2017decoupled} with a learning rate of 2e-5, a batch size of 256, and a warm-up ratio of 0.1 during training. 
Following~\citep{shi2023replug}, document embeddings are refreshed every 5k steps. 
In the case of few-shot learning, we adhere to the setup in contriever~\citep{izacard2022unsupervised}, utilizing a momentum value of 0.9995 and a temperature of 0.05, and a learning rate of 5e-5, a batch size of 512. The dense retriever is initialized with \texttt{BERT-base-uncased} model~\citep{devlin-etal-2019-bert}.
\subsection{Diverse Generation Details}
We leverage BERT~\cite{li-etal-2023-bert} to embed the questions in the pool. In order to ensure the diversity of demonstrations, we employ the classic K-Means algorithm to cluster questions based on their embeddings. The number of clusters is set to 6, and the structure of questions can be YesNo, What/When, Which, Numeric, Location, Person. We select questions by randomly choosing one question from each group. Notably, we mask entity mentions in each question to allow the clustering algorithm concentrate on the sentence structure information rather than specific instances.

\end{document}